\documentclass{article} 

\usepackage{proceed2e,times}

\usepackage{hyperref}
\usepackage{url}
\usepackage{mathtools}
\usepackage{nicefrac}
\usepackage{amsmath,amsfonts,amssymb}
\usepackage{pgfplots}
\usepackage{caption}
\usepackage{subcaption}
\usepackage[style=long,smallcaps]{glossaries}
\usepackage{verbatim}
\usepackage{psfrag}
\usepackage[off]{auto-pst-pdf}
\usepackage{xspace}
\usepackage{bbold}
\pgfplotsset{compat=newest}
\pgfplotsset{plot coordinates/math parser=false}

\usepackage{amsthm}

\newcommand{\bea}{\begin{eqnarray}}
\newcommand{\eea}{\end{eqnarray}}
\newcommand{\beq}{\begin{equation}}
\newcommand{\eeq}{\end{equation}}
\newcommand{\nn}{\nonumber}

\def\be{\begin{equation}}
\def\ee{\end{equation}}
\def\beq{\begin{eqnarray}}
\def\eeq{\end{eqnarray}}

\def\eeq{\end{eqnarray}}

\newtheorem{theorem}{Theorem}

\title{Gradient descent in Gaussian random fields as a toy model for high-dimensional optimisation in deep learning}

\author{
Mariano Chouza\\
Tower Research Capital, London \\
\And
Stephen Roberts\\
University of Oxford \\
\And
Stefan Zohren \\
University of Oxford \\
}

\begin{document}

\maketitle

\begin{abstract}
In this paper we model the loss function of high-dimensional optimization problems by a Gaussian random field, or equivalently a Gaussian process. Our aim is to study gradient descent in such loss functions or energy landscapes and compare it to results obtained from real high-dimensional optimization problems such as encountered in deep learning. In particular, we analyze the distribution of the improved loss function after a step of gradient descent, provide analytic expressions for the moments as well as prove asymptotic normality as the dimension of the parameter space becomes large. Moreover, we compare this with the expectation of the global minimum of the landscape obtained by means of the Euler characteristic of excursion sets. Besides complementing our analytical findings with numerical results from simulated Gaussian random fields, we also compare it to loss functions obtained from optimisation problems on synthetic and real data sets by proposing a ``black box'' random field toy-model for a deep neural network loss function.
\end{abstract}

\section{Introduction}

For almost a decade there have been significant advances in many areas of machine learning by applying deep learning techniques, such as in the context of image recognition  \cite{krizhevsky2012imagenet,vggnet}, generative adversarial networks \cite{goodfellow2014generative} and in reinforcement learning, most notably in the development of AlphaGo  \cite{singh2017artificial} (see also \cite{deeplearningbook} for an overview). The amount of progress, combined with some issues that were found such as robust adversarial examples \cite{goodfellow2014explaining, gilmer2018adversarial}, have led to interest in getting a better understanding of the underlying process. One contributing factor for the success of deep neural networks has to do with the nature and complexity of its loss function or energy landscape. In particular, it was found that the loss function of deep neural networks has very similar properties to random fields or Gaussian processes \cite{choromanska2015loss,lee2017deep,schoenholz2016deep}. For example, it was seen that the Hessian of such a loss function is mostly governed by the spectrum of a random matrix \cite{choromanska2015loss} which can be used to show that local minima are located in a band close to the global minimum. 

Given the above evidence that loss functions of deep neural networks share many properties with those of random energy landscapes, we want to investigate further optimization procedures in such landscapes. In particular, we model the loss function of high-dimensional optimization problem as a Gaussian random field (GRF) \cite{adler2009random}, which can also be viewed as a Gaussian Process (GP) \cite{GPbook}, and study, both theoretically as well as experimentally, the performance of gradient descent in such landscapes as well as properties of the global minimum. Recently, there has been a revived interest in studying distributional properties of GRFs within the research community working on GPs. Examples include the study of the distribution of arc length in GPs \cite{arclength}, as well as expected improvements in batch optimization \cite{batchbayesian}. We aim to fill a gap in the literature by studying distributional aspects of improvements in gradient descent in such landscapes, including proving asymptotic normality of the improved field value. Interesting scalings can be obtained by studying the optimal learning rate and comparing it with that of random search as well as the location of the global minimum both as a functions of the dimension of the parameter space.

\begin{figure}
\begin{center}
  \includegraphics[height=0.6\linewidth]{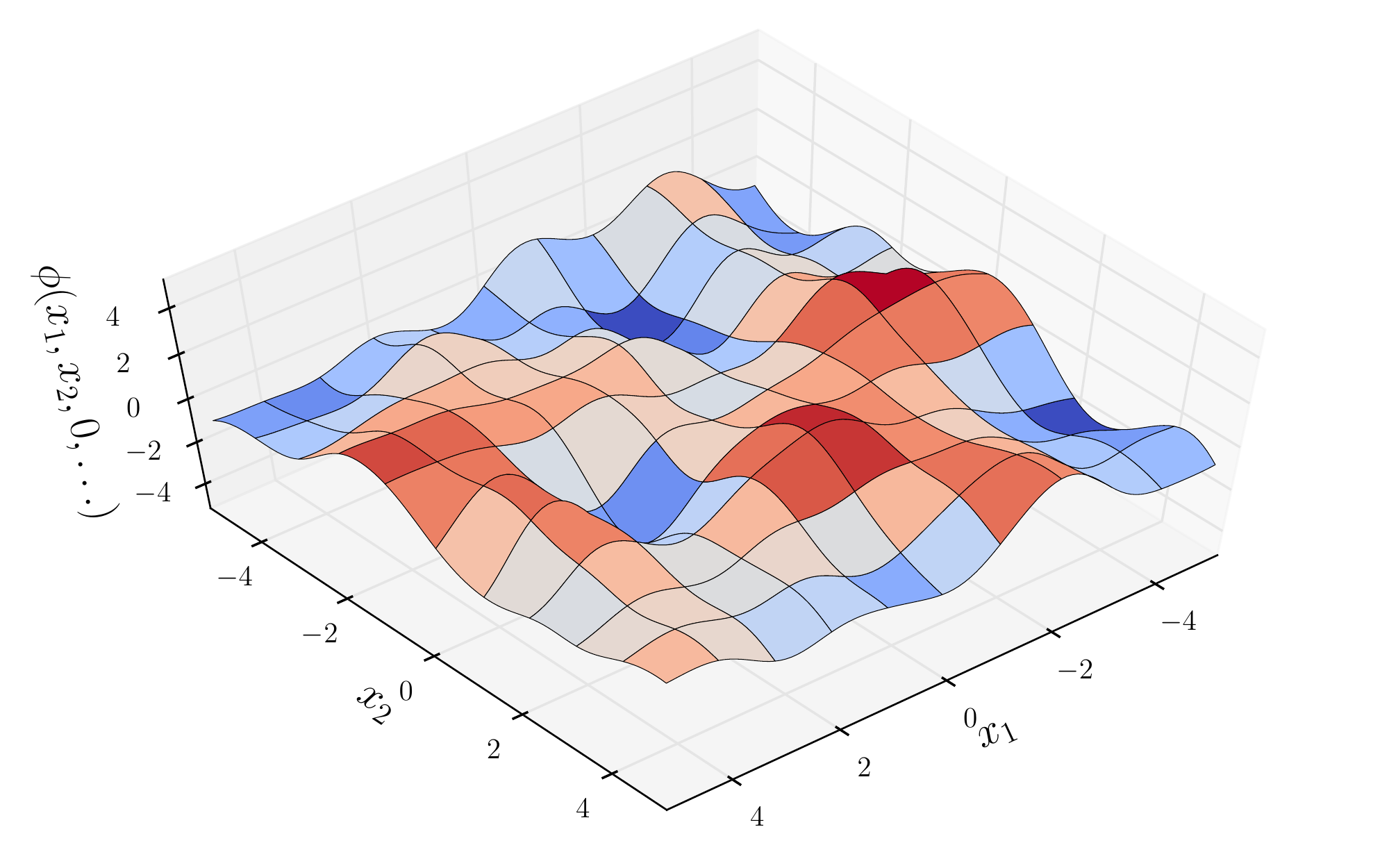}
  \caption{Two-dimensional slice of a realization in $\mathbb{R}^{500}$ of the previously defined random field.}
  \label{fig:random_field_slice}
    \end{center}
\end{figure}

As in \cite{choromanska2015loss} but differing from \cite{lee2017deep} and \cite{schoenholz2016deep}, we explicitly consider the field to be a function of the input and the parameters. Concretely, we choose the loss function to be a Gaussian random field $\phi(\mathbf{x})$ with squared exponential correlation $k(r)$ and constant mean $\mu$, where $r = \|\mathbf{x} - \mathbf{x}'\|$ is the distance between two points. As the differences between different parameters will then be just be scaling and translation, we choose $\mu = 0$ and $k(r) = \exp(-r^2/2)$ for definiteness. As an example, Figure~\ref{fig:random_field_slice} illustrates a two-dimensional slice of a 500-dimensional realization of the field.

\begin{figure}
\begin{center}
  \includegraphics[height=0.6\linewidth]{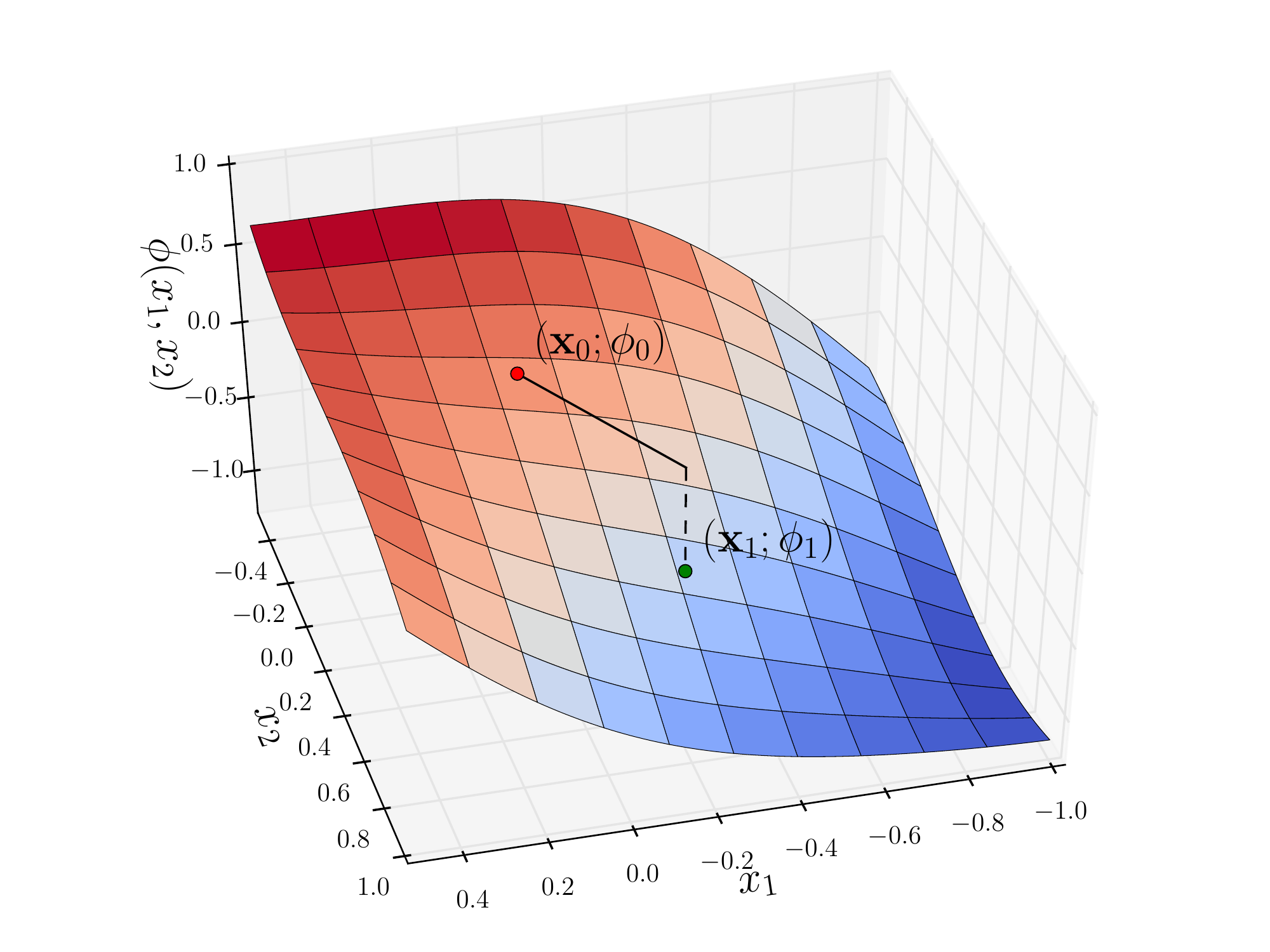}
  \caption{Gradient descent first step in a two-dimensional realization of the random field.}
  \label{fig:random_field_slice2}
  \end{center}
\end{figure}

We are concerned with analyzing properties of gradient descent in the above random energy landscape, 
\beq \label{eq:gradientdescentdef}
 \mathbf{x}_1 = \mathbf{x}_0 - \eta\nabla\phi\left(\mathbf{x}_0\right).
\eeq
Here $ \mathbf{x}_1 $ is the updated point, $\mathbf{x}_0$ is the initial point where we start our gradient descent and $\eta$ is the learning rate. As we will frequently use the values of the field and its gradient at both points, we introduce the short-hand notation $\Phi_0 \equiv \phi(\mathbf{x}_0)$, $\Phi_1 \equiv \phi(\mathbf{x}_1)$, $\boldsymbol{\Xi}_0 \equiv \nabla\phi(\mathbf{x}_0)$ and $\boldsymbol{\Xi}_1 \equiv \nabla\phi(\mathbf{x}_1)$ as illustrated in Figure~\ref{fig:random_field_slice2}.

After a short introduction to Gaussian Processes (GPs) in the next section, we present our main theoretical results in Section \ref{sectheo}. Firstly, in Section \ref{theor_mean_and_var} we obtain a formal expression for the distribution of $\Phi_1$, as well as provide analytic expressions for its expected value and variance as a function of the dimension $N$ of our parameter space. In Section \ref{opt_lr}, we use the expected value of $\Phi_1$, to compute the optimal learning rate. When comparing the optimal learning rate with that of random search, it is seen how random search gives superior results for small dimensions while gradient descent outperforms random search in larger dimensions. In Section \ref{asymptotic_normality} we prove asymptotic normality of the rescaled random variable $\Phi_1$. For most practical applications, the Gaussian approximation of the distribution of $\Phi_1$ is sufficiently close to the true distribution. In the following Section \ref{opt_comp} we compare the expected value of $\Phi_1$ with that of the global minimum in a unit ball which we estimate by means of analyzing the Euler characterise of excursion sets. The latter is found to have the same scaling with the dimension $N$ as the expected value of $\Phi_1$ but with a slightly larger per-factor. Besides the above theoretical results, we also provide numerical results and simulation experiments in Section \ref{sec:experimental}. More precisely, in Section \ref{sec:experimental:1} we numerically simulate GRFs of dimensions up to $N=500$ and verify the theoretical results obtained in the previous sections. Furthermore, we also investigate gradient descent on a toy model of GFRs which models the loss functions of deep neural networks on synthetic as well as real-life datasets and compare those with the findings on GRFs which are in good agreement.

\section{Gaussian random fields and Gaussian processes}

The Gaussian random field (GRF) we introduced in the previous section can be seen to be equivalent to saying that the field or loss function $\phi$ is given by a Gaussian Process (GP)
\beq
\phi \sim \mathrm{GP}(0,K)
\eeq
with zero mean and kernel $K( \mathbf{x}_0,  \mathbf{x}_1) = k(| \mathbf{x}_0 - \mathbf{x}_1|)$ with $k(r) = \exp(-r^2/2)$. For a comprehensive review on GPs the reader is referred to \cite{GPbook}. The GP can be understood as an infinite dimensional extension of a multivariate Gaussian distributions such that joint distributions of any finite number of points are again a multivariate Gaussians. For our problem at hand it means that the joint distribution of $\Phi_0 \equiv \phi(\mathbf{x}_0)$ and $\Phi_1 \equiv \phi(\mathbf{x}_1)$, given $\mathbf{x}_0,\mathbf{x}_1$, is Gaussian with mean zero and covariance $K( \mathbf{x}_0,  \mathbf{x}_1)$. Having a kernel which depends on the distance makes closer points more correlated where in fact the correlation goes to one if the distance goes to zero. This essentially ensures that $\phi$ will be a continuous function. This makes GPs popular choices for prior distributions over continuous functions in Bayesian statistics. One frequently occurring quantity to compute in this context is the posterior distribution of the field $\phi$ at a new point $\mathbf{x}_1$ given the value $\Phi_0$ observed at $\mathbf{x}_0$ which can be obtained by conditioning the joint distribution. The conditional distribution is indeed Gaussian, $[\Phi_1]_{cond} = \Phi_1|\Phi_0,\mathbf{x}_0,\mathbf{x}_1\sim\mathcal{N}(\mu,\Sigma)$, with conditional mean $\mu$ and conditional covariance $\Sigma$ given by,
\bea
\!\mu \!\!\!\!&=&\!\! \!\!\! K( \mathbf{x}_0,  \mathbf{x}_1)^T K( \mathbf{x}_0,  \mathbf{x}_0)^{-1} \Phi_0 \nn\\
\!\Sigma \!\!\!\!&=&\!\!\!\!\! K( \mathbf{x}_1,  \mathbf{x}_1)\! -\! K(\mathbf{x}_0,  \mathbf{x}_1)^T \! K(\mathbf{x}_0,  \mathbf{x}_0)^{-1} \! K( \mathbf{x}_0,  \mathbf{x}_1)
\eea
This is a well-known result for GPs and a similar result holds true when conditioning on more than one point. The implications of this relation are important since it essentially means that we can efficiently compute posterior updates of probability distributions at the expense of a few matrix operations.

\section{Theoretical results}
\label{sectheo}

\subsection{Distribution of the field after one step of gradient descent}
\label{theor_mean_and_var}

In the previous section we saw that the joint distribution of $( \Phi_1 , \Phi_0)^T$ is a Gaussian and that the conditional distribution can be easily obtained. Moreover, the joint distribution of the $2N+2$ dimensional vector $( \Phi_1 ,\boldsymbol{\Xi}_1, \Phi_0, \boldsymbol{\Xi}_0)^T$ is also a multivariate Gaussian with mean $\mathbf{0}$ and a covariance matrix $\Sigma$ which can easily be expressed in terms of the kernel function $k$ and its derivative,
\bea
\Sigma        &=& \left[\begin{array}{c|c}\Sigma_{11} & \Sigma_{12}\\\hline \Sigma_{21} & \Sigma_{22}\end{array}\right] \\ 
\Sigma_{11} &=& \Sigma_{22} =  \mathbb{1} \\
\Sigma_{12} &=& \Sigma_{21}^T \nonumber\\
 &=& e^{-\Delta\mathbf{x}^2/2}\left(\mathbb{1} -  \left[\begin{array}{cc} 0 & \Delta\mathbf{x}^T\\ -\Delta\mathbf{x} & \Delta\mathbf{x}\Delta\mathbf{x}^T\end{array}\right]\right)
\eea
where $\mathbb{1}\equiv \mathbb{1}_{(N+1)\times (N+1)}$ and $\Delta\mathbf{x} \equiv \mathbf{x}_1 - \mathbf{x}_0$.

The values of the field and its gradient at $\mathbf{x}_1$, $\Phi_1$ and $\boldsymbol{\Xi}_1$, are going to be Gaussian random variables conditioned on the values at $\mathbf{x}_0$, $\Phi_0$ and $\boldsymbol{\Xi}_0$, similar to the example presented in the previous section,
\beq
\left[\begin{array}{c}\Phi_1\\ \boldsymbol{\Xi}_1\end{array}\right]_{cond}\!\!\!\!\!\!\! \sim \mathcal{N}\left(\Sigma_{12}\Sigma_{22}^{-1}\! \left[\begin{array}{c}\Phi_0 \\ \boldsymbol{\Xi}_0\end{array}\right]\!, \mathbb{1} \!-\! \Sigma_{12}\Sigma_{22}^{-1}\Sigma_{21}\right).\!\!
\eeq
When multiplying out the terms and replacing $\Delta\mathbf{x} = -\eta\boldsymbol{\Xi}_0$ from \eqref{eq:gradientdescentdef}, we obtain that $\Phi_1$ follows a conditional normal distribution with mean and variance,
\bea
 m_1(\varphi_0, \xi_0^2) &:=& \mathbb{E}\left[\left.\Phi_1\right|\Phi_0=\varphi_0,\Xi_0^2=\xi_0^2\right]  \nonumber \\
 &=& e^{-\frac{\eta^2}{2}\xi_0^2}\left(\phi_0 -\eta \xi_0^2\right) \label{eq:m1}\\
v_1(\xi_0) &:=& \mathrm{Var}\left[\left.\Phi_1\right|\Phi_0=\varphi_0,\Xi_0^2=\xi_0^2\right] \nonumber\\ 
&=& 1 - e^{-\eta^2 \xi_0^2}\left(1 + \eta^2 \xi_0^2\right). \label{eq:v1}
\eea

\begin{figure}
\begin{center}
  \includegraphics[width=0.9\linewidth]{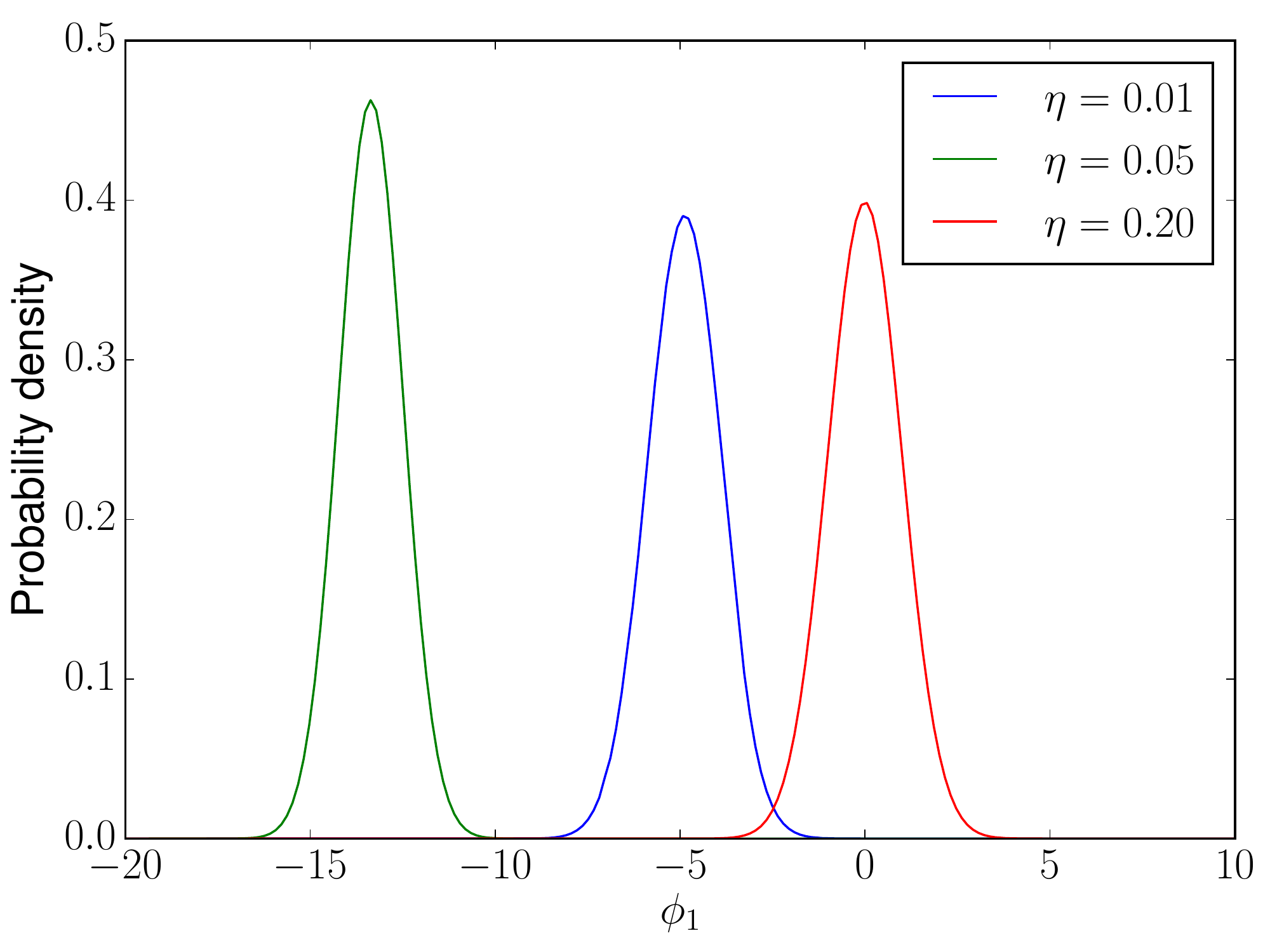}
  \caption{Numerically integrated probability density functions of $\Phi_1$ plotted for some values of $\eta$ with $N = 500$.}
  \label{fig:pdf_phi_1_n_500}
  \end{center}
\end{figure}

As $\Phi_0$ will have a $\mathcal{N}(0, 1)$ distribution and $\Xi_0^2$ will have a $\chi^2$-distribution with $N$ degrees of freedom, being the sum of the squares of $N$ independent normally distributed components, we can write the overall distribution as

\bea
 f_{\Phi_1}(\varphi_1) &=& \int_{-\infty}^{+\infty}\!\!\! d\varphi_0 \int_0^{+\infty}\!\!\! d\xi_0^2\,  \frac{  f_{\Phi_0}(\varphi_0)\,f_{\Xi_0^2}(\xi_0^2) }{\left(2\pi\,v_1(\xi_0^2)\right)^{1/2}}\times\nonumber\\
&&\times   \exp\left(-\frac{\left(\varphi_1 - m_1(\varphi_0, \xi_0^2)\right)^2}{2\,v_1(\xi_0^2)}\right),
\eea

where $f_{\Phi_0}(\varphi_0)$ is the standard normal probability density function (PDF), $f_{\Xi_0^2}(\xi_0^2)$ is the PDF for a chi-squared random variable with $N$ degrees of freedom and $m_1$, $v_1$ are the conditional mean and variance of $\Phi_1$ as given in \eqref{eq:m1}-\eqref{eq:v1}. For illustration purposes, we plotted the resulting PDF by means of numerical integration as shown in Figure~\ref{fig:pdf_phi_1_n_500}.

A closed form expression for the distribution of $\Phi_1$ seems out of reach, however, we can calculate its moments as well as show asymptotic normality later. For now, we focus on the first moments which can be easily derived, 
\bea
\mathbb{E}[\Phi_1] &=&\mathbb{E}_{\varphi_0, \xi_0^2}[ m_1(\varphi_0, \xi_0^2)] = -\eta \, \mathbb{E}_{\xi_0^2}\left[ \xi_0^2 e^{-\frac{\eta^2}{2}\xi_0^2}  \right] \nn\\
&=& -N\eta\left(\eta^2 + 1\right)^{-N/2-1},
\eea
where the last expectation value is computed by integrating over the PDF of the $\chi^2$-distribution. Similarly, for the variance one obtains
\bea
\mathrm{Var}(\Phi_1) &=&  \mathbb{E}_{\xi_0^2}[ v_1(\xi_0) ] \nn\\
&=& 1 -  \mathbb{E}_{\xi_0^2}\left[ e^{-\eta^2 \xi_0^2}\left(1 + \eta^2 \xi_0^2\right)  \right] \nn\\
&=& 1 + N\eta^2\left(1 + 2\eta^2\right)^{-\frac{N}{2}-2}\left(N + 1 - 2\eta^2\right) \nonumber \\
&& - N^2\eta^2\left(\eta^2 + 1\right)^{-N-2}.
\eea

It can be observed that the mean and variance of $\Phi_1$ converge to those of $\Phi_0$ in the limit where $\eta\to 0$, as we stay in the same point and also when $\eta\to+\infty$, as the gradient only gives significant information in a neighborhood of $\mathbf{x}_0$ of size $\mathcal{O}(1)$. 

\subsection{Optimal learning rate and comparison to random search}
\label{opt_lr}

The optimal learning rate $\eta_{opt}$ can be easily derived by computing $\frac{d}{d\eta} \mathbb{E}[\Phi_1] |_{\eta = \eta_{opt}}= 0$, yielding 
\beq 
\eta_{opt} = (N + 1)^{-\frac{1}{2}}
\eeq
Computing the second derivative shows that indeed it is a minimum. We can now obtain  the expected value of the field for the optimal learning rate which is given by
\bea
 \mathbb{E}\left[\left.\Phi_1\right|\eta = \eta_{opt}\right] &=& -N(N + 1)^{-\frac{1}{2}}\left(\frac{N+2}{N + 1}\right)^{-\frac{N}{2}-1}  \nn\\
&=& -\sqrt{\frac{N}{e}} + \mathcal{O}\left(N^{-\frac{1}{2}}\right).
 \eea

\begin{figure}
\begin{center}
  \includegraphics[width=0.9\linewidth]{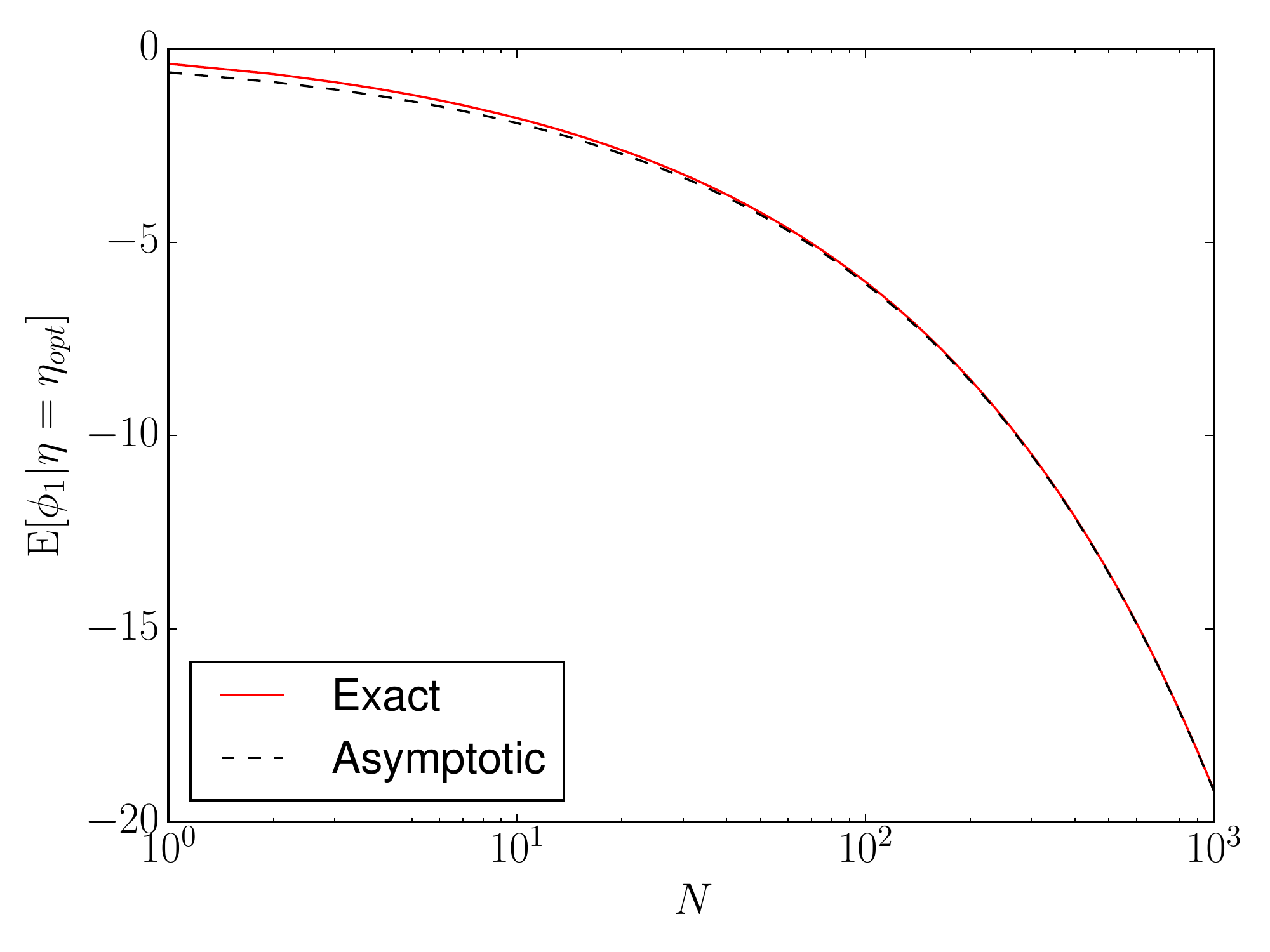}
  \caption{Expected value for $\Phi_1$ as a function of $N$ when using the optimal learning rate.}
  \label{fig:mean_phi_1_opt_eta}
  \end{center}
\end{figure}

The expected value of $\Phi_1$ will improve as the square root of the number of dimensions $N$, as shown in Figure~\ref{fig:mean_phi_1_opt_eta}, and, as we show in Section \ref{opt_comp}, this is within a constant factor of the minimum field value within a unit radius ball. We would not expect significantly better results, as the information provided by the gradient decays very fast for distances greater than the correlation length (1 in our case). 

The expected step length will also tend to 1 for large values of $N$ when using the optimal learning rate. That can be seen by using the previously discussed fact that the squared gradient has a $\chi^2$-distribution with $N$ degrees of freedom and computing the expectation:

\bea
\mathbb{E}[\eta_{opt}\Xi_0] &=& \frac{1}{\sqrt{N+1}}\sqrt{2}\frac{\Gamma(N/2+1/2)}{\Gamma(N/2)} \nn\\
&=& 1 + \mathcal{O}(N^{-1}).
\eea

Taking $N = 500$, a single step of gradient descent with the optimal learning rate gives us an expected value of
\beq
\mathbb{E}[\Phi_1|\eta=\eta_{opt},N=500]  \approx -\sqrt{\frac{500}{e}} \approx -13.56.
\eeq
To put this value into context we compare it to random search. Since any evaluation of the random field would give a value smaller than the above with probability $F_{\mathcal{N}}(-13.56) \approx 3.46\cdot 10^{-42}$, where $F_{\mathcal{N}}$ is the cumulative distribution function of a standard normal, more than $10^{41}$ tries would be needed on average to get to a value smaller than that from random search. On the other hand, for $N = 1$ the expected value after a gradient descent step would only be
\beq
\mathbb{E}\left[\left.\Phi_1\right|\eta=\eta_{opt},N=1\right] = -\frac{2}{3\sqrt{3}} \approx -0.385.
\eeq
This value or better would be obtained by random search in an average of $F_{\mathcal{N}}(-0.385)^{-1} \approx 2.85$ tries. The difference exemplifies how gradient descent becomes increasingly powerful when moving to higher dimensional optimization. Below, in Section~\ref{opt_comp}, we further investigate how this compares to the value of the global minimum.

\subsection{Asymptotic normality of the distribution of the field}
\label{asymptotic_normality}

We now analyze convergence of the distribution of $\Phi_1$ for $N\to \infty$ when we scale the learning rate around its optimal value, namely, under the scaling
\beq \label{scalingLR}
\eta = \frac{X}{\sqrt{N}},
\eeq
where $X$ is the rescaled learning rate. Under this scaling we see that the expected value of $\Phi_1$
\beq
\mathbb{E}[\Phi_1] = \mu_N(X) +..., \quad  \mu_N(X):= - \sqrt{N} X e^{-\frac{X^2}{2}}
\eeq
is of $\mathcal{O}(N^{1/2})$ while the variance
\beq \label{defsigma2}
\mathrm{Var}(\Phi_1) = \sigma^2(X) + ... , \quad \sigma^2(X):= 1+X^2 e^{-X^2}
\eeq
remains finite. We will now show that
\begin{theorem}
As the dimension $N\to\infty$, the rescaled field value after a single step of gradient descent converges asymptotically to a normal distribution:
\beq
\Phi_1 - \mu_N(X) \xrightarrow{d} \mathcal{N} (0, \sigma^2(X) )
\eeq
where $X$, $\mu_N(X)$, $\sigma^2(X)$, are defined in \eqref{scalingLR} - \eqref{defsigma2}.
\end{theorem}

\begin{figure}
\begin{center}
  \includegraphics[width=0.9\linewidth]{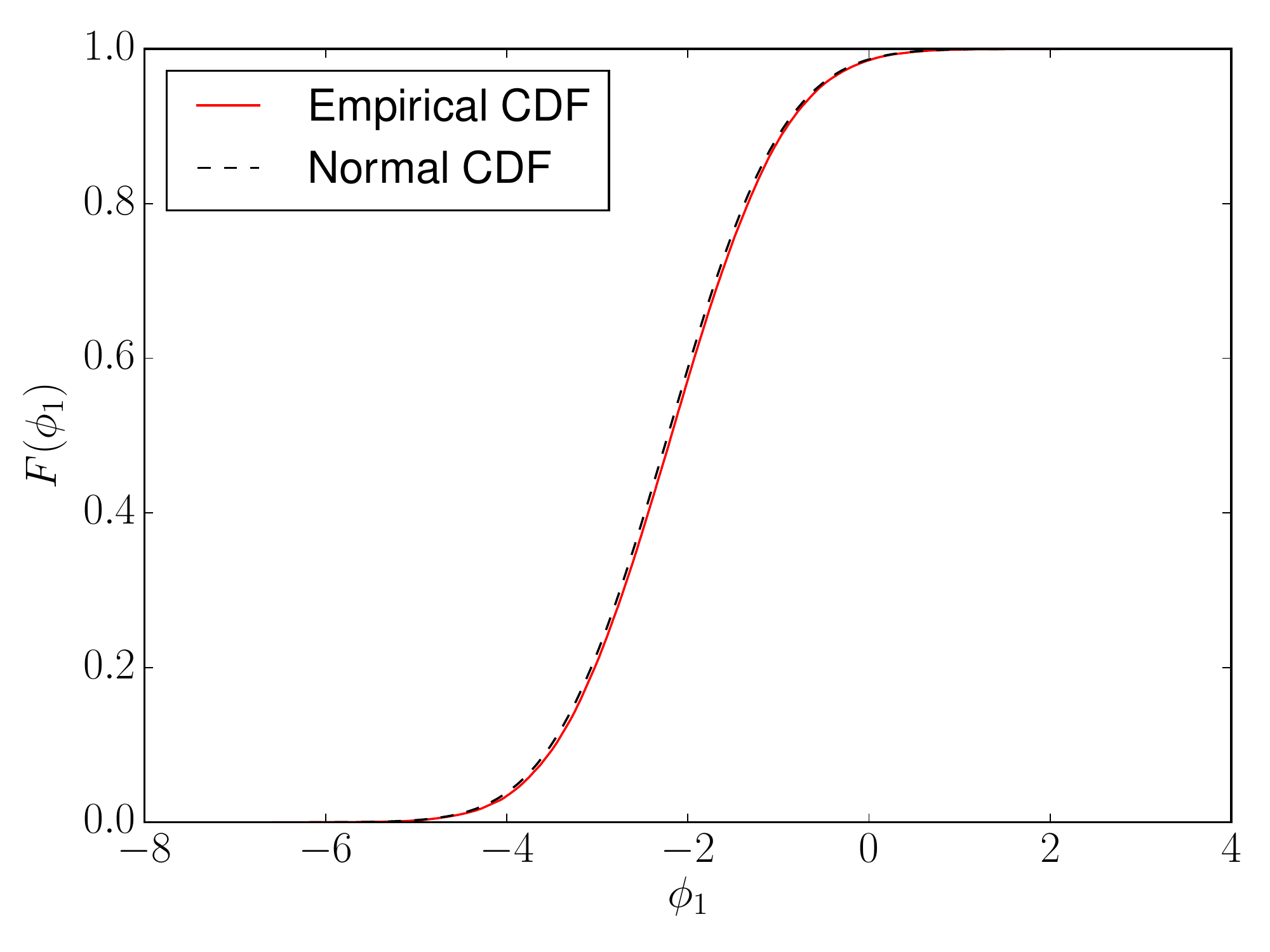}
  \caption{Comparison of the empirical cumulative distribution function of simulated values of $\phi_1$ with the normal cumulative distribution function with corresponding mean and standard deviation ($\eta = 0.1\,\eta_{opt}$). }
  \label{fig:pdf_phi_1_n_500_exp}
  \end{center}
\end{figure}

\begin{proof}
To prove the theorem we first compute the moment generating function 
\bea
\mathbb{E}\left[e^{t\Phi_1}\right] \!\!\! &=& \!\!\!  \mathbb{E}_{\varphi_0, \xi_0^2} \!\!   \left[ \exp\left\{t \, m_1(\varphi_0, \xi_0^2)  + \frac{t^2}{2} \, v_1(\xi_0)  \right\} \right] \nn \\ 
 \!\!\!  &=& \!\!\!   e^{\frac{t^2}{2}} \mathbb{E}_{\xi_0^2} \!\!   \left[ \exp\left\{  - \frac{t^2}{2} \eta^2 \xi_0^2 e^{-\eta^2\xi_0^2} + \right.  \right.  \nn \\
  \!\!\!  &&    \,\,\, \quad \quad \left.\left. - \,\, t \eta \xi_0^2 e^{-\eta^2\xi_0^2/2}  \right\}  \right]
\eea
Inserting the scaling relation for the learning rate, \eqref{scalingLR}, and using a saddle point expansion of the integral over $\xi_0^2$ when writing out the expectation one can see that the above expression is given to leading order by
\bea
\mathbb{E}\left[e^{t\Phi_1}\right] &=&  e^{\frac{t^2}{2}}  \exp\left\{   - \frac{t^2}{2} X^2  e^{-X^2}  +\right. \nn \\  
&& \quad \quad   \quad  \left.\, \, \, \, \,- \, t X \sqrt{N} e^{-X^2/2}   \right\} \,+ ... \nn \\ 
&=&  \exp\left\{ \frac{\sigma^2(X) \, t^2}{2} + t \mu_N(X)  \right\}  + ...
\eea
where the expectation over $\xi_0^2$ collapsed to its saddle point which to leading order is given by $\mathbb{E}\xi_0^2 = N$. Thus
\beq
\lim_{N\to\infty} \mathbb{E}\left[e^{t(\Phi_1 - \mu_N(X)) }\right] = e^{\frac{\sigma^2(X) \, t^2}{2} }
\eeq
which proves the above convergence. 
\end{proof}

Figure \ref{fig:pdf_phi_1_n_500_exp} shows an example of the normal approximation of the distribution of $\Phi_1$ for finite $N$. Details of the numerical analysis will be presented in Section \ref{sec:experimental}.

\subsection{Comparison with optimal values}
\label{opt_comp}

In the previous sections we have analyzed the expected value of the field after a step of gradient descent. A natural follow-up  is to ask how does it compare with the global extremum of the field. As we are working with a Gaussian random field with a covariance that decays to zero, we can expect to find values with arbitrarily large magnitude at enough distance, but a more useful comparison can be done by restricting ourselves to a unit ball $\mathcal{B}_N(\mathbf{x}_0)$ around the random starting point $\mathbf{x}_0$.

There is no known analytical expression for the expected value of a Gaussian random field maximum or minimum in any multidimensional domain, but a number of powerful estimation techniques have been developed \cite{adler2009random, aldous2013probability, azais2009level}. Most of them are based on finding quantities that are related to the extrema and computing their expectations. In our case we will use the exact computation of the expected Euler characteristic, as described in \cite{adler2009random}, to get an estimate for the number of connected components of an excursion set and use that estimate to get the expected value of the minimum.

\begin{figure}
\begin{center}
  \includegraphics[width=\linewidth]{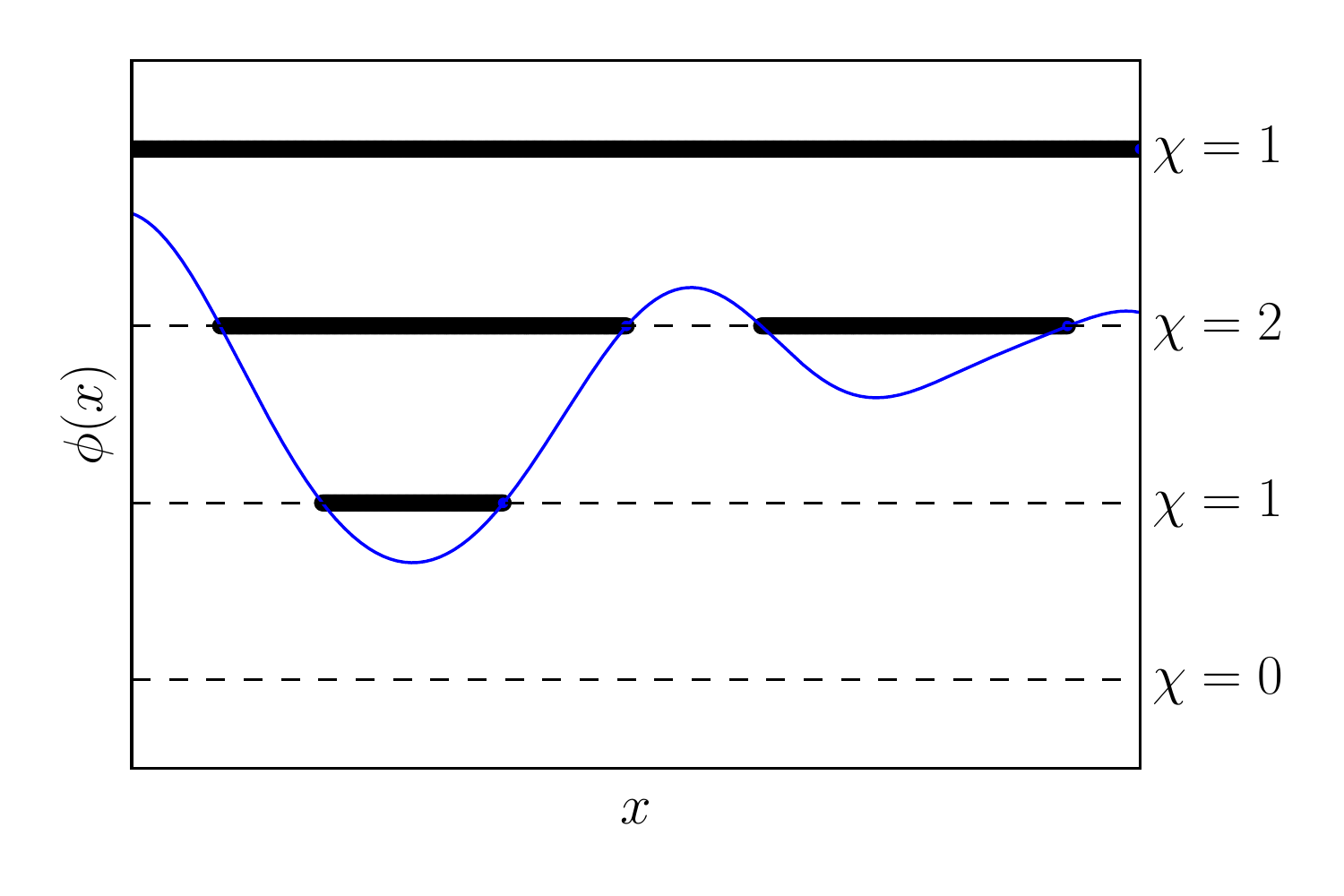}
  \caption{Excursion sets $A_u$ and their Euler characteristics for different values of $u$.}
  \label{fig:excursion_sets}
  \end{center}
\end{figure}

In general, the Euler characteristic can be seen as the unique functional $\chi$ from a family of subsets $\mathcal{A}$ of a manifold $\mathcal{M}$ to the integers that has the following properties:
\begin{itemize}
\item $\chi(\emptyset) = 0$.
\item $\chi(X) = 1$ if $X$ is contractible.
\item $\chi(X \cup Y) = \chi(X) + \chi(Y)$ if $X \cap Y = \emptyset$.
\end{itemize}
In the following discussion we will only use these properties of the Euler characteristic and assume that all the sets under consideration are included in $\mathcal{A}$. A detailed proof of the uniqueness of the Euler characteristic and a description of the family $\mathcal{A}$ in the context of random fields can be found in \cite{adler2009random}.

In our case the manifold is the unit ball, $\mathcal{M}=\mathcal{B}_N(\mathbf{x}_0)$. Now we can define an excursion set $A_u$ in $\mathcal{B}_N(\mathbf{x}_0)$ as the subset of $\mathcal{B}(\mathbf{x}_0)$ composed by the points where the field $\varphi$ reaches a value of $u$ or smaller. It is clear then that $A_u$ will be non-empty if and only if $u > \min_{\mathbf{x}\in\mathcal{B}(\mathbf{x}_0)}\phi(\mathbf{x})$, allowing us to connect geometrical properties of the excursion set $A_u$ with the value of the minimum.

As our random field is continuous, if we start $u$ from a large positive value and gradually decrease it, we expect the excursion set $A_u$ to start being all of $\mathcal{B}(\mathbf{x}_0)$, then getting some holes, being disconnected, turning into a few contractible components and finally ending as the empty set, as shown in Figure~\ref{fig:excursion_sets}. If our excursion set has the form of disjoint contractible components, its Euler characteristic gives us the number of components  and we will be able to use its expected value for different values of $u$ to estimate the value of the minimum.

Following \cite{adler2009random}, we note that it has been proved that the expected Euler characteristic $\chi$ of the excursion set $A_u$ under the conditions we described is given by
\beq
\mathbb{E}\left[\chi(A_u)\right] = \sum_{j=0}^N \mathcal{L}_j(\mathcal{B}_N(\mathbf{x}_0))\rho_j(u),
\eeq
where $\mathcal{L}_j(\mathcal{B}_N(\mathbf{x}_0))$ are the Lipschitz-Killing curvatures of $\mathcal{B}_N(\mathbf{x}_0)$ which are given by
\beq
\mathcal{L}_j(\mathcal{B}_N(\mathbf{x}_0)) = \binom{N}{j}\frac{\omega_N}{\omega_{N-j}}, \quad \omega_j = \frac{\pi^{j/2}}{\Gamma(j/2+1)}
\eeq
and $\rho_j(u)$ is given by
\beq
\rho_j(u) = (2\pi)^{-(j+1)/2}H_{j-1}(u)e^{-\frac{{u}^2}{2}},
\eeq
with $H_j$ being the Hermite polynomials.

As shown in Figure~\ref{fig:excursion_sets}, we anticipate the expected Euler characteristic starting at 1 for large positive values of $u$ (the whole set has characteristic 1), to be 1 for values of $u$ that leave a single non-empty excursion set with high probability (a small ``droplet" has characteristic 1) and to decrease to 0 when it starts being less probable to find a non-empty excursion set (in other words, when $u$ is below the minimum).

\begin{figure}
  \includegraphics[width=\linewidth]{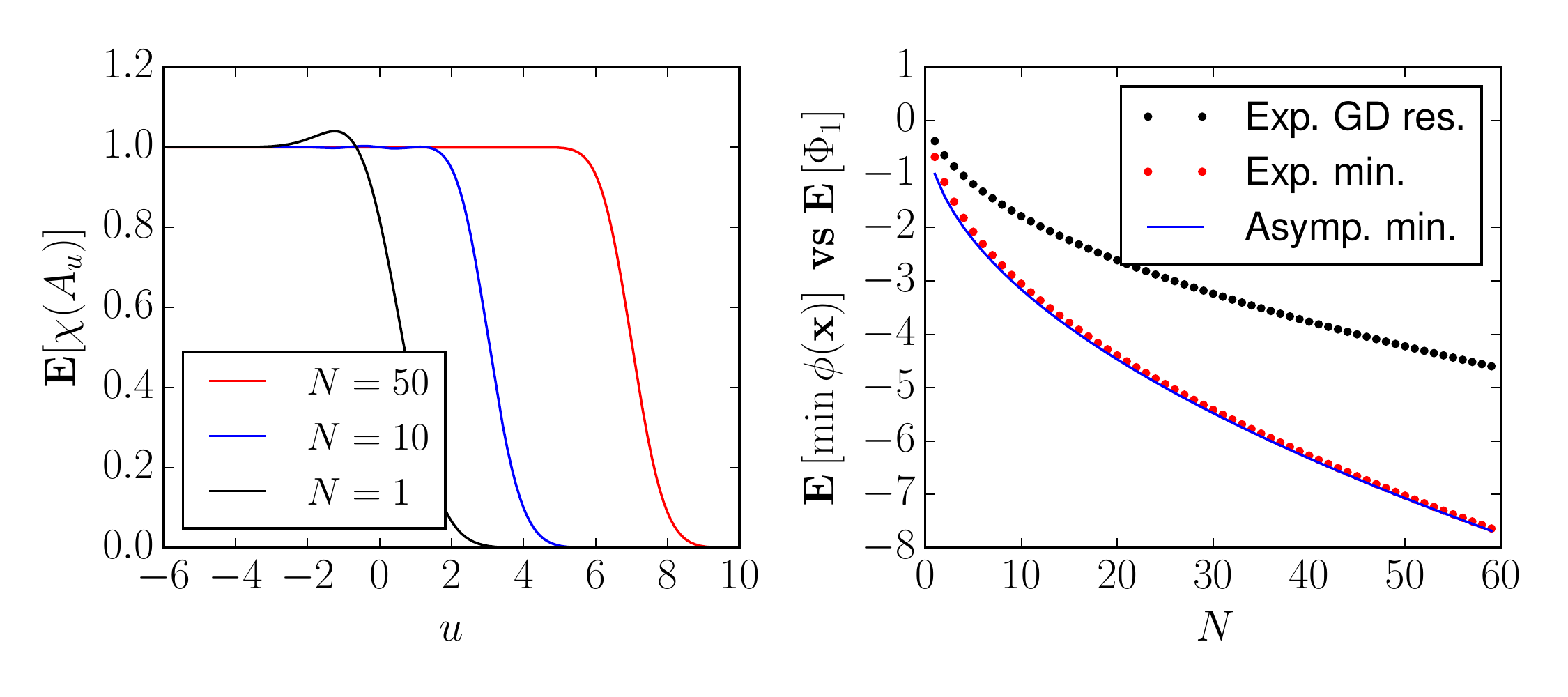}
  \caption{The left figure shows the shape of the expected Euler characteristic as a function of the threshold level $u$ for multiple values of $N$. Those Euler characteristic values are used to estimate the expected values of the minimum (Exp. min.) and its asymptotics (Asymp. min.), shown in the right figure compared against the results of a gradient descent step (Exp. GD res.), both as a function of $N$.}
  \label{fig:extrema}
\end{figure}

The expected behavior can be seen in Figure~\ref{fig:extrema}, with the threshold increasing in absolute value as we move to higher dimensional spaces. If we estimate the expected minimum as the value where the expected Euler characteristic is 0.5, we find its values are well approximated by $-\sqrt{N}$, i.e.

\beq
\mathbb{E}\left[\min_{\mathbf{x}\in\mathcal{B}(\mathbf{x}_0)}\phi(\mathbf{x})\right] \approx - \sqrt{N}.
\eeq

Comparing those values with the values of the field after a gradient descent step, as found in Section~\ref{opt_lr}, we see they differ by a constant factor of $\sqrt{e}$.

Applying this bound to get the expected value of the minima inside the unit ball for $N = 500$, we find it will be $\mathbb{E}[\min \phi] \approx-\sqrt{500}\approx-22.36$, which should be contrasted with the expected field value after one step of gradient descent, $\mathbb{E}[ \Phi_1] \approx-13.56$, as obtained in Section~\ref{opt_lr}.

\section{Experimental results}
\label{sec:experimental}

\subsection{Random field simulation}
\label{sec:experimental:1}

\begin{figure}
  \includegraphics[width=\linewidth]{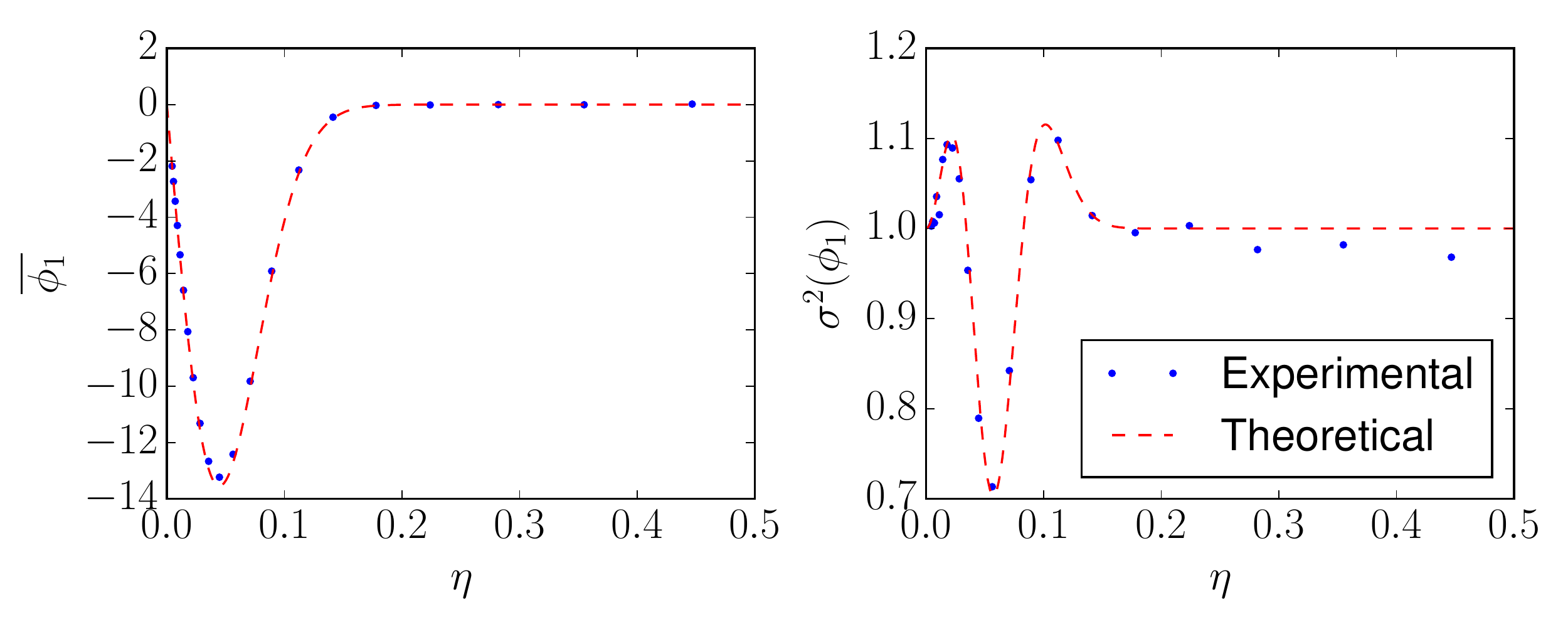}
  \caption{Comparison of sample mean and variance with the theoretically expected values for multiple values of the learning rate $\eta$ and $N = 500$.}
  \label{fig:mean_var_phi_1_n_500}
\end{figure}

Our experiments will require generating approximate instances of Gaussian random fields in spaces of high dimensionality, with values of $N$ reaching 500. Most of the conventional methods for simulating random fields \cite{bertschinger2001multiscale, lang2011fast} don't scale well to a large number of dimensions, as they generate explicit grids representing the field values.

We can take the spectral representation \cite{adler2009random} of the random field,
\beq
\phi(\mathbf{x}) = \int e^{i \mathbf{z}^T \mathbf{x}} W(d\mathbf{z}),
\eeq
that can be seen as the Fourier transform of the field expressed as a stochastic integral, and use Monte Carlo sampling to approximate the integration over $\mathbf{z}$. In that way, we obtain an approximate instance of the random field expressed as the real part of the sum of $M$ complex exponentials
\beq
 \phi_{sim}(\mathbf{x}) = \Re\,\mathbf{w}^T\overline{\exp}\left(i\,Z\mathbf{x}\right),
\eeq
where $\overline{\exp}$ is component-wise exponentiation, $\mathbf{w} \sim \mathcal{CN}(\mathbf{0}_M, M^{-1}\mathbf{1}_M)$ is a complex Gaussian random vector and $Z \in \mathbb{R}^{M \times N}$ is a real Gaussian random matrix with independently distributed elements $Z_{mn} \sim \mathcal{N}(0, 1)$. This can be considered a multidimensional variant of the randomization method described in \cite{kramer2007comparative}, although in a high dimensional context it is important to ensure that the number of samples $M$ is significantly higher than the number of dimensions $N$ to avoid confining the gradient to a low dimensional subspace.

The gradient can then be computed by differentiating the previous expression:
\beq
  \nabla\phi_{sim}(\mathbf{x}) = -\Im\,Z^T\,\mathrm{diag}\left(\mathbf{w}\right)\overline{\exp}\left(i\,Z\mathbf{x}\right).
\eeq

The value of $M$, being the number of samples, will determine how accurate our representation of the random field will be. Higher values will increase the amount of computational resources required and lower values will produce a lower quality realization of the random field. A value of $2\cdot 10^4$ was found to give high quality results for $N \le 500$ at acceptable computational cost.

We first start by comparing the expected values for the sample mean and variance computed in section~\ref{theor_mean_and_var} with experimental results. With the previously discussed representations and for 20 different values of the learning rate $\eta$, we can do one step of gradient descent for $10^4$ different starting points distributed uniformly in $[-10^6, 10^6]^{500}$. The resulting sample means and variances are shown in Figure~\ref{fig:mean_var_phi_1_n_500} compared with the theoretical expectations and we can see they match them quite accurately.

To compare the expected distribution with the empirical one, we repeated the gradient descent step simulation using $10^5$ points and $\eta = 0.1\,\eta_{opt}$. The simulated results can be seen in Figure~\ref{fig:pdf_phi_1_n_500_exp} and they also fit very closely with the expected distribution.

\subsection{Experiments on synthetic and real datasets}
\label{sec:experimental:2}

In this section we show how this random field model can be used to classify real data. To do that, we introduce a toy model in which we take a standard multilayer network based binary classifier and we replace the entire network by a ``black box'' loss function, given by a static random field, with no adjustable internal parameters. The $N_P$-dimensional parameter vector $\boldsymbol{\beta}$, replacing the weights of a normal network, and the $i$-th $N_I$-dimensional input to be classified $\mathbf{x}_{\rm input}^i$ are concatenated to get a $N$-dimensional vector, with $N=N_P+N_I$,
\beq
\mathbf{x}^i = \left[\begin{array}{c}\boldsymbol{\beta}\\ \mathbf{x}_{\rm input}^i\end{array}\right],
\eeq
that is the random field input.

It is a normal practice \cite{deeplearningbook} in classifier networks to use softmax as the activation function in the last layer and cross-entropy as the loss function. As we are replacing the rest of the network with a random field and using only two classes, the output of the classifier can be written as
\beq
y^i = \mathrm{sigmoid}\left(\phi\left(\mathbf{x}^i\right)\right),
\eeq
where $\mathbf{x}^i$ is the input vector associated with the $i$-th input instance and $\mathrm{sigmoid}(z)=1/(1+\exp(-z))$ is the sigmoid function.

As usual in supervised binary classification problems, we associate a true class label $y^i_{\rm true} \in \{0, 1\}$ to each of our input instances $\mathbf{x}_{\rm input}^i$ and we try to minimize the cross entropy loss between the true labels $y^i_{\rm true}$ and the classifier outputs $y^i$, i.e.\ $L_i = y^i_{\rm true} \log y^i - (1 - y^i_{\rm true}) \log (1 - y^i).$


\begin{figure}
  \includegraphics[width=\linewidth]{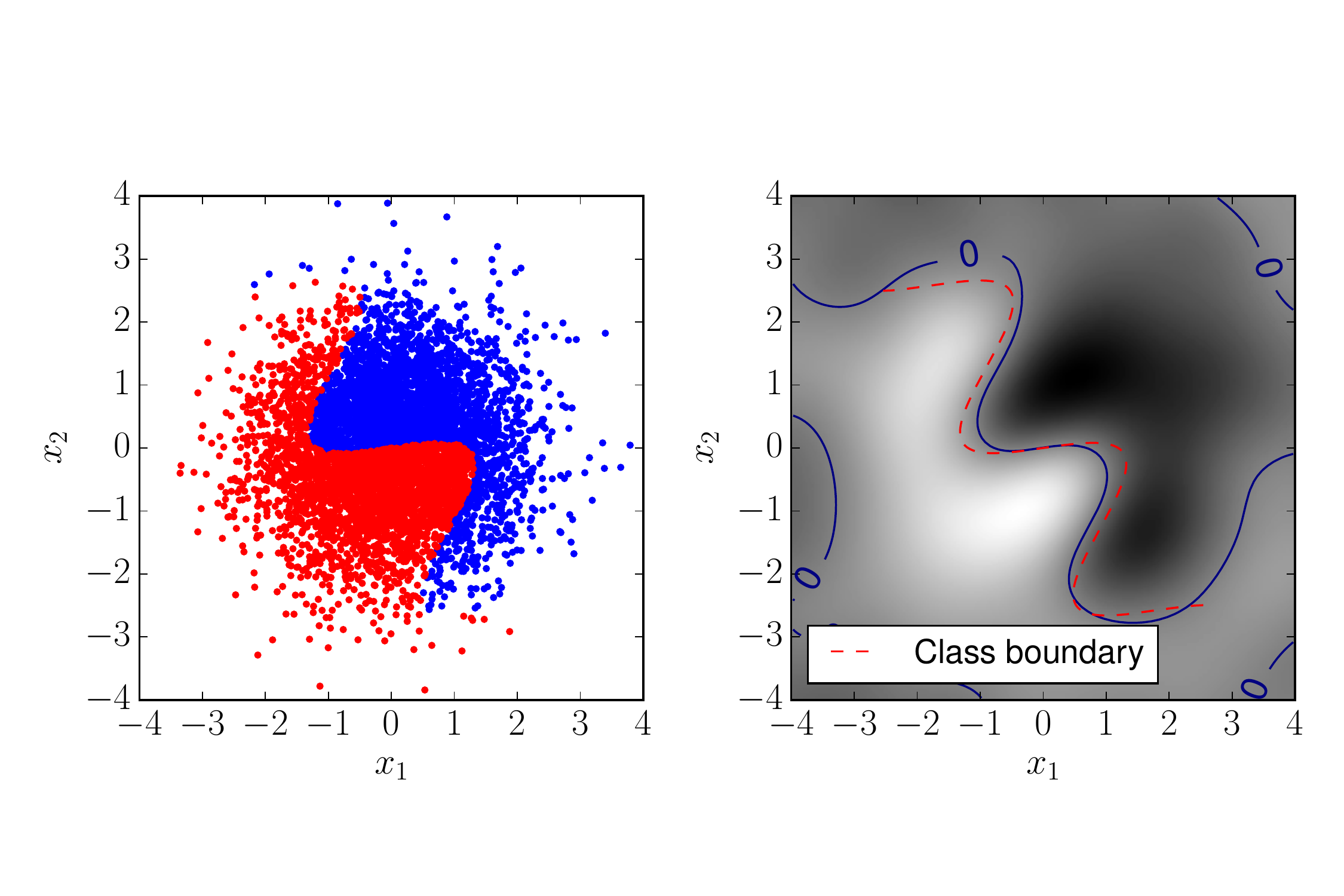}
  \caption{Original dataset and associated values of the random field $\phi$ for the trained parameters, showing how the classification works.}
  \label{fig:sine_field}
\end{figure}

The training process is standard minibatch gradient descent. By analogy with neural networks, where only the weights are updated, only the parameter vector $\boldsymbol{\beta}$ is updated after each minibatch. Following usual practice, we divide the  input data into training and test sets, using the training set to select a value for  the parameter vector $\boldsymbol{\beta}$ and the test set to evaluate the accuracy of the classifier. Furthermore, both sets of input instances are normalized to mean 0 and mean norm 1, matching the scale of the data set distribution to the correlation scale of the random field. This is empirically observed to make a significant difference in classification accuracy.

One way of visualizing the training process is to think of the parameter vector $\boldsymbol{\beta}$ as selecting a random field slice. Then the gradient descent over the loss function will try to select a slice where the naive Bayes  decision surface divides both classes, putting the instances where $y_{true}^i = 1$ in the positive side and the instances with $y_{true}^i = 0$ in the negative side of the surface. This process can be seen clearly in Figure~\ref{fig:sine_field}, where the parameter vector after training can be seen as selecting a 2D slice of the random field $\phi$, where the intersection of the naive Bayes decision boundary with the slice is close to the class boundary.

Note that the here proposed toy model of a ``black box'' random field that mimics the energy landscape of a deep neural network is different to GP classification \cite{GPclassification,GPbook} or other kernel methods such as support vector machines \cite{svm,kernelbook} in the sense that we combine the input and parameter vector into a joint input vector to a GRF which is kept fixed during the learning process.

To test the classification power of these random field instances, we run the training and test process  over two simple data sets:

\begin{itemize}
\item Normally distributed points in the $\mathbb{R}^2$ plane separated by a sine function (see Figure~\ref{fig:sine_field}), with 6000 elements in the training set and 1000 in the test set ($N_I = 2$, as we are talking about points in the plane).
\item MNIST \cite{MNIST0,MNIST}, modified to classify the digits as even or odd and using 60000 elements in the training set and 10000 in the test set ($N_I = 784$ in this case).
\end{itemize}

The training is done using a fixed batch size of 128 and 10 epochs, combined with different learning rates and values of $N$ to evaluate their impact over test set accuracy.

\begin{figure}
  \includegraphics[width=\linewidth]{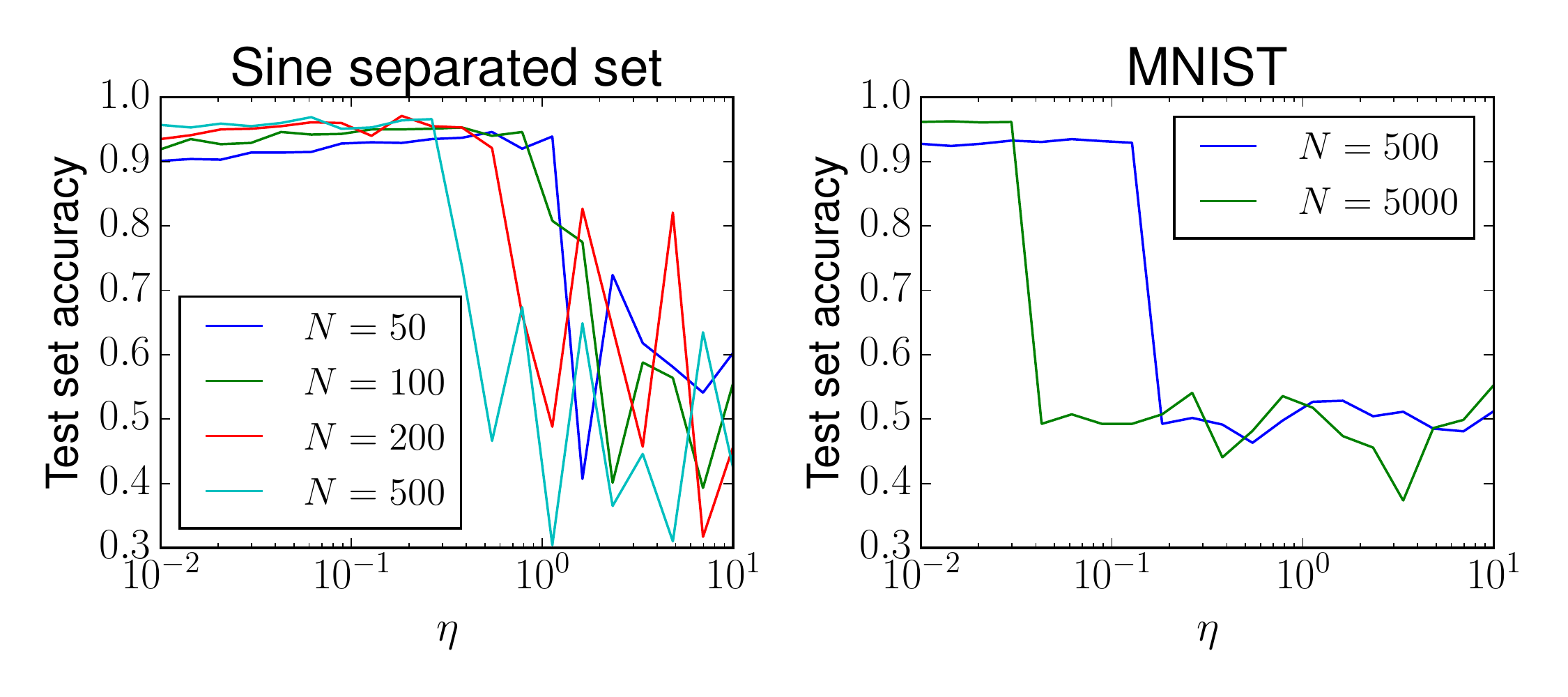}
  \caption{Sine separated and MNIST test set accuracy as a function of the learning rate $\eta$ and the dimensionality $N$.}
  \label{fig:sine_mnist}
\end{figure}

The test set accuracy for MNIST with $N = 5000$ is over 96\% showing that, even though it is far from matching the state of the art, the model has significant classification power. When compared with models with a similar number of parameters, the model is competitive \cite{MNIST}.

We can observe in Figure~\ref{fig:sine_mnist} that accuracy doesn't depend on the learning rate until reaching a critical value and then it drops to random performance. The MNIST drop for $N = 500$ is at $\eta \approx 0.13$ and for $N = 5000$ at $\eta \approx 0.03$,

\beq
0.03 \cdot \sqrt{5000} \approx 2.12 \approx 2.68 \approx 0.13 \cdot \sqrt{500},
\eeq

roughly matching the scaling found before in the single step regime.

\section{Conclusion}

The successes of deep learning as well as some unexpected weaknesses, such as the difficulty of combining good generalization and resistance to adversarial examples, have led to a significant research effort aiming to understand why their training process performs so well in high dimensional problems. The complex structure of deep neural networks error landscapes makes it difficult to understand how the optimization process is working, but observing its performance in a simple random field model can help to clarify some of the reasons behind its successes and limitations.

In this work we aim to get a better understanding of gradient descent as a tool for high dimensional optimization by obtaining theoretical and empirical results about its performance over Gaussian random fields. Following a brief introduction, we establish some asymptotic results about the distribution of field values reached after a single step of gradient descent. Those results are then compared  with a theoretical estimate of the extreme field values at a similar distance. Finally, we compare the previously obtained theoretical results with experimental simulations, while also showing that the ``black box'' Gaussian random field model is capable of solving realistic classification tasks.

We show our theoretical results about the distribution of values after a gradient descent step in Section~\ref{sectheo}. Starting in Section~\ref{theor_mean_and_var}, we obtain the first and second moments as a function of the learning rate $\eta$ and the number of dimensions $N$ of the parameter space. In the following Section~\ref{opt_lr} we use the previously derived expressions to get the optimal value for the learning rate as a function of the number of dimensions, finding that $\eta_{\rm opt}(N) \approx N^{-1/2}$ for large values of $N$. Using that optimal learning rate we show that the expected value of the field after a gradient descent step is approximately $\mathbb{E}[\Phi_1]\approx -(N/e)^{1/2}$, comparing  very favorably with the values that can be obtained through a random search when $N \gg 1$, as those are independent of the dimensionality of the space. Closing our analysis of the distribution of values, we prove in Section~\ref{asymptotic_normality} that in the high dimensional limit the distribution of the values after the gradient descent step is approximately normal, with a variance that is independent of the dimensionality and a mean that is proportional to $ -N^{1/2}$. Finally, in Section~\ref{opt_comp} we show using the expected Euler characteristic of excursion sets that the expected minimum inside the unit ball will only differ from the expected value we obtain through one step of gradient descent with the optimal learning rate by a factor of $\sqrt{e}$ in the $N \gg 1$ limit.

In Section~\ref{sec:experimental} we start by showing how we simulate a high-dimensional random field and comparing the experimental gradient descent results with the previous theoretical results in Section~\ref{sec:experimental:1}, finding them to be in good agreement. Finally, we show that the model we obtain by replacing a neural network by a Gaussian random field can be trained by gradient descent, obtaining competitive results in a simple synthetic dataset and in MNIST, once we take into account that the model is only using 5000 parameters.

The introduced ``black box'' GRF model is successful at combining nontrivial classification performance in realistic datasets with being simple enough to be susceptible to exact theoretical analysis. A possible line of future investigation would be to look at other aspect of deep neural networks through the lens of our toy model such as the interpretability of hidden layer neurons in image classification tasks or transfer learning. That could be combined with extending these results to the normal multistep minibatch training process.


\begin{thebibliography}{10}

\bibitem{krizhevsky2012imagenet}
Alex Krizhevsky, Ilya Sutskever, and Geoffrey~E Hinton.
\newblock Imagenet classification with deep convolutional neural networks.
\newblock In {\em Advances in neural information processing systems}, pages
  1097--1105, 2012.

\bibitem{vggnet}
K.~Simonyan and A.~Zisserman.
\newblock Very deep convolutional networks for large-scale image recognition.
\newblock In {\em International Conference on Learning Representations}, 2015.

\bibitem{goodfellow2014generative}
Ian Goodfellow, Jean Pouget-Abadie, Mehdi Mirza, Bing Xu, David Warde-Farley,
  Sherjil Ozair, Aaron Courville, and Yoshua Bengio.
\newblock Generative adversarial nets.
\newblock In {\em Advances in neural information processing systems}, pages
  2672--2680, 2014.

\bibitem{singh2017artificial}
Satinder Singh, Andy Okun, and Andrew Jackson.
\newblock Artificial intelligence: Learning to play go from scratch.
\newblock {\em Nature}, 550(7676):550336a, 2017.

\bibitem{deeplearningbook}
Ian Goodfellow, Yoshua Bengio, and Aaron Courville.
\newblock {\em Deep Learning}.
\newblock MIT Press, 2016.
\newblock \url{http://www.deeplearningbook.org}.

\bibitem{goodfellow2014explaining}
Ian~J Goodfellow, Jonathon Shlens, and Christian Szegedy.
\newblock Explaining and harnessing adversarial examples.
\newblock \url{http://arxiv.org/abs/1412.6572}, 2014.

\bibitem{gilmer2018adversarial}
Justin Gilmer, Luke Metz, Fartash Faghri, Samuel~S Schoenholz, Maithra Raghu,
  Martin Wattenberg, and Ian Goodfellow.
\newblock Adversarial spheres.
\newblock \url{http://arxiv.org/abs/1801.02774}, 2018.

\bibitem{choromanska2015loss}
Anna Choromanska, Mikael Henaff, Michael Mathieu, G{\'e}rard~Ben Arous, and
  Yann LeCun.
\newblock The loss surfaces of multilayer networks.
\newblock In {\em Artificial Intelligence and Statistics}, pages 192--204,
  2015.

\bibitem{lee2017deep}
Jaehoon Lee, Yasaman Bahri, Roman Novak, Samuel~S Schoenholz, Jeffrey
  Pennington, and Jascha Sohl-Dickstein.
\newblock Deep neural networks as {G}aussian processes.
\newblock \url{http://arxiv.org/abs/1711.00165}, 2017.

\bibitem{schoenholz2016deep}
Samuel~S Schoenholz, Justin Gilmer, Surya Ganguli, and Jascha Sohl-Dickstein.
\newblock Deep information propagation.
\newblock \url{http://arxiv.org/abs/1611.01232}, 2016.

\bibitem{adler2009random}
Robert~J Adler and Jonathan~E Taylor.
\newblock {\em Random fields and geometry}.
\newblock Springer Science \& Business Media, 2009.

\bibitem{GPbook}
Carl~Edward Rasmussen and Christopher K.~I. Williams.
\newblock {\em Gaussian Processes for Machine Learning (Adaptive Computation
  and Machine Learning)}.
\newblock The MIT Press, 2005.

\bibitem{arclength}
Justin~D. Bewsher, Alessandra Tosi, Michael~A. Osborne, and Stephen~J. Roberts.
\newblock Distribution of {G}aussian process arc lengths.
\newblock In {\em Proceedings of the 20th International Conference on
  Artificial Intelligence and Statistics (AISTATS)}, 2017.

\bibitem{batchbayesian}
Nikitas Rontsis, Michael~A. Osborne, and Paul~J. Goulart.
\newblock Distributionally ambiguous optimization techniques for batch
  {B}ayesian optimization.
\newblock \url{http://arxiv.org/abs/1707.04191}, 2017.

\bibitem{aldous2013probability}
David Aldous.
\newblock {\em Probability approximations via the {P}oisson clumping
  heuristic}, volume~77.
\newblock Springer Science \& Business Media, 2013.

\bibitem{azais2009level}
Jean-Marc Aza{\"\i}s and Mario Wschebor.
\newblock {\em Level sets and extrema of random processes and fields}.
\newblock John Wiley \& Sons, 2009.

\bibitem{bertschinger2001multiscale}
Edmund Bertschinger.
\newblock Multiscale {G}aussian random fields and their application to
  cosmological simulations.
\newblock {\em The Astrophysical Journal Supplement Series}, 137(1):1, 2001.

\bibitem{lang2011fast}
Annika Lang and J{\"u}rgen Potthoff.
\newblock Fast simulation of {G}aussian random fields.
\newblock {\em Monte Carlo Methods and Applications}, 17(3):195--214, 2011.

\bibitem{kramer2007comparative}
Peter~R Kramer, Orazgeldi Kurbanmuradov, and Karl Sabelfeld.
\newblock Comparative analysis of multiscale {G}aussian random field simulation
  algorithms.
\newblock {\em Journal of Computational Physics}, 226(1):897--924, 2007.

\bibitem{GPclassification}
C.~K.~I. Williams and D.~Barber.
\newblock Bayesian classification with {G}aussian processes.
\newblock {\em IEEE Transactions on Pattern Analysis and Machine Intelligence},
  20(12):1342, 1998.

\bibitem{svm}
Bernhard~E. Boser, Isabelle~M. Guyon, and Vladimir~N. Vapnik.
\newblock A training algorithm for optimal margin classifiers.
\newblock In {\em Proceedings of the fifth annual workshop on Computational
  learning theory -- COLT '92}, page 144, 1992.

\bibitem{kernelbook}
Bernhard Sch{\"o}lkopf and Alexander~J. Smola.
\newblock {\em Learning with Kernels}.
\newblock MIT Press, 2002.

\bibitem{MNIST0}
Y.~LeCun, L.~Bottou, Y.~Bengio, and P.~Haffner.
\newblock Gradient-based learning applied to document recognition.
\newblock {\em Proceedings of the IEEE}, 86(11):2278, 1998.

\bibitem{MNIST}
Yann LeCun, Corinna Cortes, and Christopher~J.C. Burges.
\newblock The {MNIST} database.
\newblock \url{http://yann.lecun.com/exdb/mnist/}.

\end{thebibliography}

\end{document}